\documentclass[runningheads]{llncs}

\usepackage[mobile]{eccv}

\usepackage{eccvabbrv}

\usepackage{graphicx}
\usepackage{booktabs}
\usepackage{lipsum}
\usepackage{multirow}
\usepackage{tikz}
\usepackage{csquotes}

\usepackage[accsupp]{axessibility}  

\usepackage[pagebackref,breaklinks,colorlinks,citecolor=eccvblue]{hyperref}

\usepackage{orcidlink}

\begin{document}

\title{NamedCurves: Learned Image Enhancement via Color Naming} 

\titlerunning{NamedCurves: Learned Image Enhancement via Color Naming}

\author{David Serrano-Lozano\inst{1,2} \and
Luis Herranz\inst{3} \and Michael S. Brown\inst{4} \and \newline
Javier Vazquez-Corral\inst{1,2}}

\authorrunning{D. Serrano-Lozano et al.}

\institute{Computer Vision Center, Barcelona, Spain \and Universitat Autònoma de Barcelona, Barcelona, Spain \and Universidad Autónoma de Madrid, Madrid, Spain \and York University, Toronto, Canada\\
\email{\{dserrano,jvazquez\}@cvc.uab.cat}, \email{luis.herranz@uam.es}, \email{mbrown@eecs.yorku.ca}\\
\url{namedcurves.github.io}}

\maketitle

\begin{figure}[th]
  \centering
  \vspace{-6mm}
    \includegraphics[width=\linewidth]{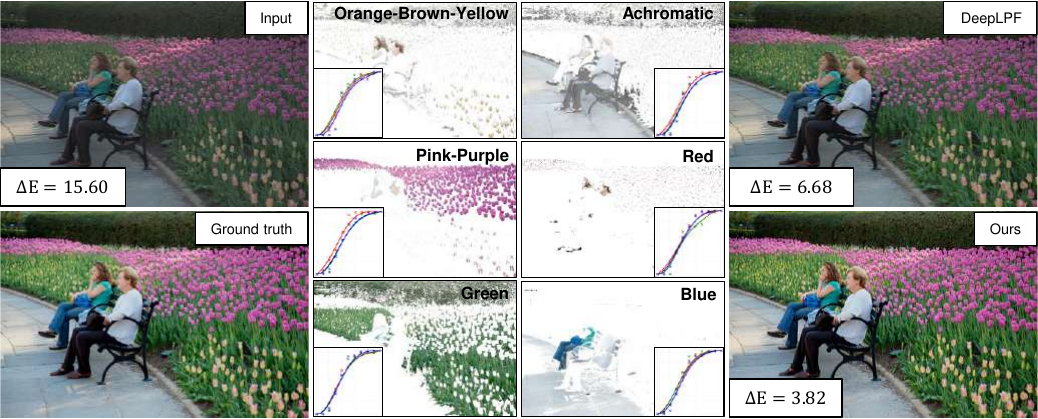}
    \caption{Column 1 displays an input image corrected by a photo-editing expert (denoted as {\it ground truth}). Our proposed method decomposes the image based on color naming and learns a tone-curve correction to mimic the expert's style (shown in columns 2-3). Results comparing the input, our results, and the approach by \cite{moran2020deeplpf} are reported in terms of the color distance metric $\Delta E_{00}$.}
    \vspace{-9mm}
  \label{fig:teaser}
\end{figure}

\begin{abstract}
A popular method for enhancing images involves learning the style of a professional photo editor using pairs of training images comprised of the original input with the editor-enhanced version. When manipulating images, many editing tools offer a feature that allows the user to manipulate a limited selection of familiar colors.  Editing by color name allows easy adjustment of elements like the "blue" of the sky or the "green" of trees. Inspired by this approach to color manipulation, we propose NamedCurves,  a learning-based image enhancement technique that separates the image into a small set of named colors. Our method learns to globally adjust the image for each specific named color via tone curves and then combines the images using an attention-based fusion mechanism to mimic spatial editing. We demonstrate the effectiveness of our method against several competing methods on the well-known Adobe 5K dataset and the PPR10K dataset, showing notable improvements.
\end{abstract}

\section{Introduction}
\label{sec:intro}

Color plays a vital role in photography, enhancing focal points, evoking emotions, and enriching storytelling. 
Whether through vibrant hues or subtle tones, understanding the importance of colors is crucial for photographers seeking to evoke specific responses.  Despite significant advancements in camera technology, amateurs and professionals still often resort to post-capture image enhancement to enhance an image's quality. However, manual enhancement can be challenging for those lacking expertise, time, or a well-developed aesthetic sense.

A potential solution to avoid manual adjustment is to learn a deep network model that can mimic the image editing style of a skilled photographer or colorist. These methods leverage a dataset of image pairs with the original and corresponding artist-edited images. It is interesting to consider the tools provided to the artists for performing the image editing. Many photo editing software applications (e.g., Adobe Photoshop\cite{photoshop}) provide users with the ability to manipulate the image based on a small set of fixed colors (e.g., red, green, yellow, orange, blue, purple).  Interestingly, the predefined colors selected by software tools are similar to those linguists have found to be universal across languages~\cite{berlin1991basic}, a research topic often referred to as {\it color naming}.

\paragraph{\textbf{Contribution:}}~We propose to leverage the use of color naming decomposition for image enhancement. In particular, we introduce NamedCurves, a learning-based image enhancement method that decomposes images into color names and estimates a tone curve in the form of smooth, differential Bezier curves (see Figure \ref{fig:teaser}). This is followed by an attention-based fusion scheme that combines the images modified by the individual color curves, simulating local image editing. We compare our method with several state-of-the-art image enhancement methods on the MIT-Adobe-5K and PPR10K datasets. Our color naming scheme  outperforms competing methods in terms of PSNR and $\Delta E$. 

\section{Related Work}
\label{sec:related}
Related works are discussed for color naming and data-driven image-based enhancement methods that model professional editing styles. 

\subsection{Color Naming}\label{subsec:colornaming-relatedwork}

Color naming is crucial for product designing, photography, and vision research \cite{xue2023integrating, regier2007color, szafir2017modeling, bahng2018coloring}. Berlin and Kay \cite{berlin1991basic} conducted a study on the basic color lexicon across various languages and discovered universal semantics. Their seminal analysis showed that the evolution of basic color vocabularies is influenced by visual physiology, which limits the possible composite categories to a small number of those. The 11 color names found that most societies and cultures share are: \textit{orange}, \textit{brown}, \textit{yellow}, \textit{white}, \textit{grey}, \textit{black}, \textit{pink}, \textit{purple}, \textit{red}, \textit{green} and \textit{blue}. 

\begin{figure}[t]
    \centering
    \includegraphics[width=\linewidth]{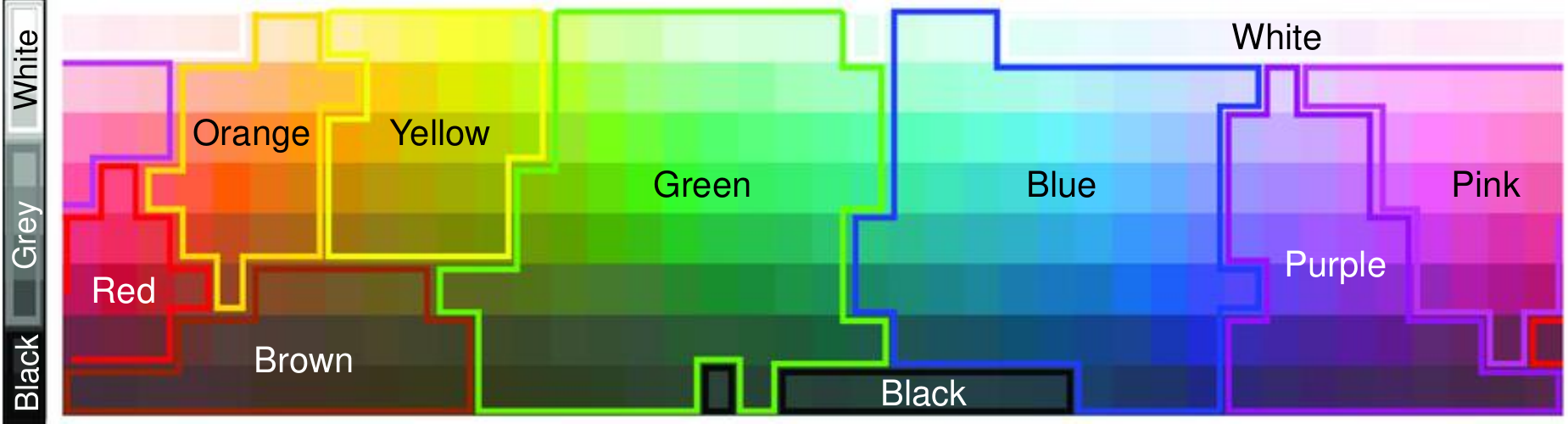}
    \caption{Van de Weijer et al. \cite{van2009learning} color names grouped in the Munsell color array. The color names are \textit{orange}, \textit{brown}, \textit{yellow}, \textit{white}, \textit{grey}, \textit{black}, \textit{pink}, \textit{purple}, \textit{red}, \textit{green} and \textit{blue}.}
    \label{fig:munsell-color-array}
\end{figure}

Following Berlin and Kay's research, different studies (e.g., \cite{van2009learning, benavente2008parametric, yu2018weakly})  aimed at predicting the boundaries between each of the color names. For example, Figure~\ref{fig:munsell-color-array} shows the standard Munsell color array using Van de Weijer et al.~\cite{van2009learning} color classification based on color naming. 

\begin{figure}[t]
    \centering
    \includegraphics[width=\linewidth]{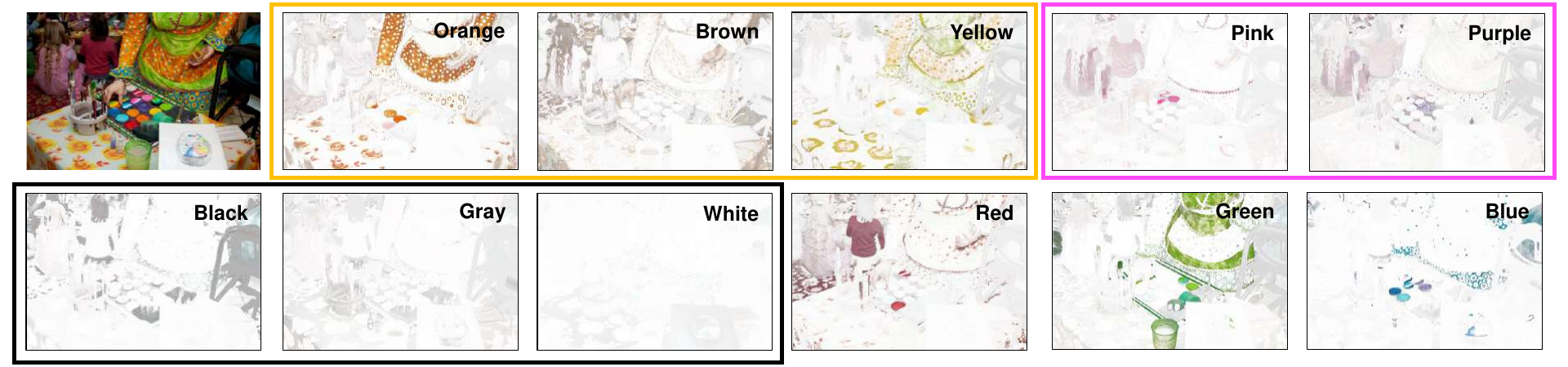}
    \caption{Van de Weijer et al. \cite{van2009learning} color naming method applied pixel-wise to the top-left image. The other 11 images show the 11 probability color names maps. Each color is displayed with a different map to aid visualization. Note that some linguistic color names share approximately the same hue and only differ in intensity--- e.g., \textit{pink} and \textit{purple}. As tone curves are defined for all the intensity range we group: \textit{orange-brown-yellow}, \textit{pink-purple}, and \textit{white-grey-black}. This grouping is represented by the boxes.}
    \label{fig:color-naming-image}
\end{figure}

These methods work in the following manner. Given an RGB value in the sRGB color space, color naming methods produce an 11-d vector that corresponds to the probability of the RGB value belonging to each of the specific color names listed above. This is visualized in Figure \ref{fig:color-naming-image}, where we show an original image and the 11-probability maps coded with a color map to aid visualization. As described in Section \ref{subsec:colornaming-method}, our method leverages the Van de Weijer et al. \cite{van2009learning} color naming strategy to decompose each image in basic colors. However, we combine colors with similar hues (e.g., brown and orange) resulting in six color maps. In Figure \ref{fig:color-naming-image} the colors grouped are shown in boxes.

\subsection{Learned Image Enhancement}\label{subsec:imageenhancement}
The need to provide users with tools to allow easy image enhancement has grown significantly due to the ease of photo-taking with smartphones.  Initially, histogram equalization was a primary method for enhancing contrast in images \cite{stark2000adaptive, reza2004realization}. Subsequently, local operators \cite{aubry2014fast, durand2002fast} and color correction techniques based on color constancy \cite{van2007edge} were introduced. Since the introduction of the MIT-Adobe-5K dataset by Bychkovsky et al. \cite{bychkovsky2011learning}, which contains 5,000 images retouched by 5 experts, data-driven methods have emerged as one of the preferred means to improve image quality. 

One category of these data-driven methods involves estimating intermediate or physical parameters for image retouching. Guo et al. \cite{guo2020zero} proposed Zero-DCE, the first method to formulate low-light image enhancement as a curve estimation problem. Their deep network estimates pixel-wise curves to modify the dynamic range of input images. This groundbreaking work influenced subsequent methods, such as CURL \cite{moran2021curl}, which estimates piecewise linear curves for HSV, RGB, and CIELab color spaces; FlexiCurve \cite{li2023flexicurve}, which estimates sets of piecewise curves and blends them via a Transformer and LTMNet \cite{zhao2022learning} that learns a grid of tone curves to locally enhance an image. Additionally, Moran et al. \cite{moran2020deeplpf} introduced DeepLPF, inspired by Photoshop's local filters tool, which estimates elliptical, gradual, and polynomial filters for local image editing. Wang et al. \cite{wang2019underexposed} proposed an alternative approach, estimating intermediate illumination maps for under-exposed images instead of directly learning image-to-image mappings.

Lookup tables (LUTs) represent another widely used method for image manipulation, typically manually tuned and fixed in camera imaging pipelines or photo editing tools. Zeng et al. \cite{zeng2020learning} proposed 3DLUT, a method to learn these 3D LUTs from annotated data with a small convolutional neural network. Building upon this, Yang et al. \cite{yang2022adaint} proposed AdaInt, a mechanism to achieve a more flexible sampling by learning the non-uniform sampling intervals. Wang et al. \cite{wang2021real} also presented a modification of 3DLUT that incorporates spatial information to compute the image transformation.

Conversely, image-to-image methods estimate directly a mapping to modify the input images without intermediate steps. Generative adversarial networks (GANs) are frequently employed for such tasks. Chen et al. \cite{chen2018deep}, Ni et al. \cite{ni2020towards}, and Jiang et al. \cite{jiang2021enlightengan} proposed unpaired learning schemes using single GANs to estimate enhanced versions of input images directly.

As in previous methods~\cite{moran2021curl, li2023flexicurve, zhao2022learning}, we use tone curves to manipulate images. However, we propose to leverage the use of color naming decomposition and an attention-based fusion scheme to mimic the image editing style of an expert. 

\section{Proposed Method}

\begin{figure}[t]
  \centering
   \includegraphics[width=\linewidth]{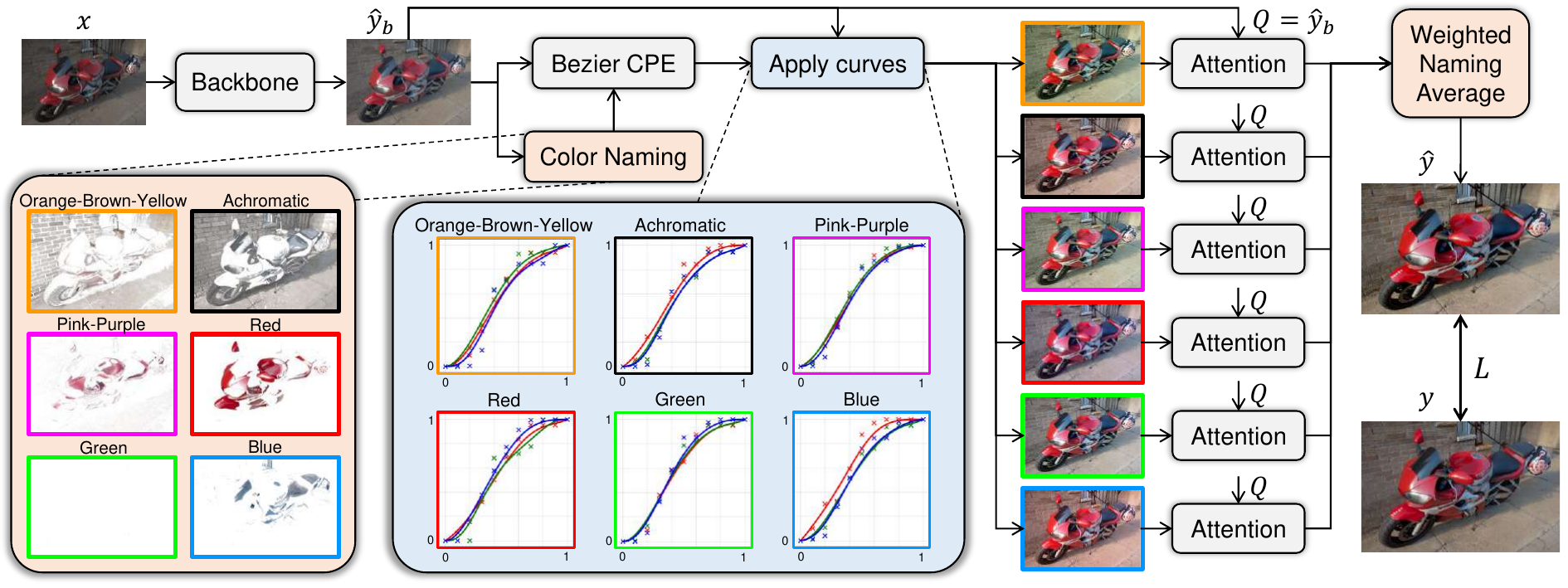}
  \caption{Overview of the proposed method, NamedCurves. Our method aims to enhance an input image $x$. First, we use a UNet-like backbone to standardize the input image into a canonical latent space. Next, we decompose the standardized image $\hat{y}_b$ into six color probability maps (shown color-coded in the figure to aid visualization). Next, we learn a set of Bezier curves for each color name to manipulate the standardized image $\hat{y}_b$, obtaining six distinct globally adjusted images. Finally, an attention mechanism is used to fuse the edited images using as \textit{Query} the standardized image $\hat{y}_b$ and as \textit{Key} and \textit{Value} the corresponding edited image. Our learning-based method uses an objective function $L$ to compare the expert-retouched image with our final result $\hat{y}$.}
  \label{fig:model-overview}
\end{figure}

Figure \ref{fig:model-overview} shows an overview of our proposed method, NamedCurves.  Our method aims to enhance a low-quality RGB input image $x$, by a learned model that outputs an enhanced version $\hat{y}$. This enhanced image is derived as close as possible to the expert-retouched image $y$, based on some objective function $L$.  

Our method consists of four main components, which are detailed in the following sections, including the loss function used for optimizing the framework. The approach first applies a DNN backbone that standardizes the input image into a canonical latent space. Next, we use color naming to decompose the image into six color maps. After color naming decomposition, a neural network learns a set of Bezier tone curves to manipulate each color map globally. Finally, an attention mechanism combines the edited images to achieve local editing effects.

\subsection{Backbone}\label{subsec:backbone}

One challenge faced by learning-based image enhancement methods is that input images, $x$, can be captured using different cameras with different settings and under different lighting conditions. This may impact our ability for consistent color naming.  Similar to the method by Moran et al.~\cite{moran2020deeplpf}, we use a UNet-like backbone to standardize the input images. 

Our backbone is inspired by the LPIENet \cite{conde2023perceptual} architecture. We use MobileNet layers (\texttt{Conv-DWConv-eLU}) \cite{howard2017mobilenets} and a CBAM module \cite{woo2018cbam}---a combination of spatial and channel attention. The backbone consists of three encoder blocks and two decoder blocks connected by multi-resolution skip connections. Each encoder block consists of the following: two MobileNet layers, a CBAM attention block, and a max-pooling layer. The decoder blocks follow the same structure except for the max-pooling layers that are replaced by bilinear upsampling layers. The multi-resolution skip connections consist of three parallel branches of convolutional layers with different dilation rates. Two of the paths consist of two \texttt{Conv-LeakyReLU} blocks to extract local information, while the other path consists of three \texttt{Conv-LeakyReLU-MaxPooling} blocks, an \texttt{AveragePooling} and a \texttt{LinearLayer} to extract global information. As in \cite{moran2021curl, marnerides2018expandnet, gharbi2017deep}, we found that skip connections at different resolutions improve the performance against backbones with simple skip connections. 

\subsection{Color Naming}\label{subsec:colornaming-method}

We aim to decompose the standardized image $\hat{y}_b$ into a set of likelihood colors maps to focus different branches of the model. Due to the importance of memory colors---the green of the grass or the blue of the sky--- in aesthetics \cite{shing2010influences,topfer2000quantitative} we decided to use Color Naming, a perceptually-based color decomposition.

We used the color naming model from Van de Weijer et al. \cite{van2009learning} to obtain the probability maps for each color name. This model inputs an sRGB color value and outputs the probability of this color to belong to each of the $11$ color naming categories, namely \textit{red, blue, green, yellow, pink, purple, orange, brown, white, grey, black}. When applied to an image, the model operates for each pixel, which returns a set of probability maps.

We note that some linguistic color names share similar hues, but only differ in intensity. For example, orange and brown or pink and purple. As tone curves are defined for all the intensity ranges, it will be beneficial to group these colors together. To this end, we reduce the set of $11$ probability maps to just $6$ by grouping \textit{orange-brown-yellow}, \textit{pink-purple}, and \textit{white-grey-black} (referring to this last one as \textit{achromatic}). The combined map for these cases is just the addition of the individual maps, and therefore, they are still probabilities (the sum of all the maps for a specific pixel is $1$)---see supplementary for further details.

Figure~\ref{fig:model-overview} shows the assignment of an input image to the six color maps. Note that the images have been color-coded to help visualize their probabilities, however, the colors associated with these maps are the original RGB values from the input (see supplemental materials). In the following section, we describe how each color map conditions a different set of RGB tone curves.

\subsection{Bezier Curve Estimation}\label{subsec:Beziercontrolpointsestimator}

Similar to prior works \cite{guo2020zero, moran2021curl, li2023flexicurve}, we leverage tone curves to remap the shadows, midtones, and highlights of the image $\hat{y}_b$ conditioned by the color naming probability maps. Tone curves represent global adjustments of the intensity from the input level to the output level. The curves are applied pixel-wise in each color channel as 1D Look-Up Tables. Bennet and Finlayson \cite{bennett2023simplifying} demonstrated that tone adjustments are typically simple curves for a large dataset of enhanced images or can be well-approximated as such. We use Bezier curve parametrization to generate smooth and continuous tone curves from discrete control points.

Specifically, we aim to estimate one global curve for each RGB channel $c$ and color name $n$. Each curve is parameterized by $M$ control points. We evenly distribute these control points along the input axis, with the first control point fixed at $(0, 0)$. Consequently, we only need to estimate the control points' output axis values instead of the two point coordinates, resulting in $M\!-\!1$ parameters per curve. Thus, the Bezier formulation $B^{n, c}$ of a curve can be expressed as:
\begin{equation}
    B^{n, c}(i) = \sum_{m=0}^{M\!-\!1} P_m^{n, c} \binom{M\!-\!1}{m} (1-i)^{(M-1-m)} i^{m},
    \label{eq:Bezier}
\end{equation}
where $i \in [0, 1] $ is an image channel pixel, $P_m^{n, c}$ denotes the $m$-th control point for the color name $n$ and channel $c$, and $M$ is the total number of control points. 

\begin{figure}[t!]
    \centering
    \includegraphics[width=\linewidth]{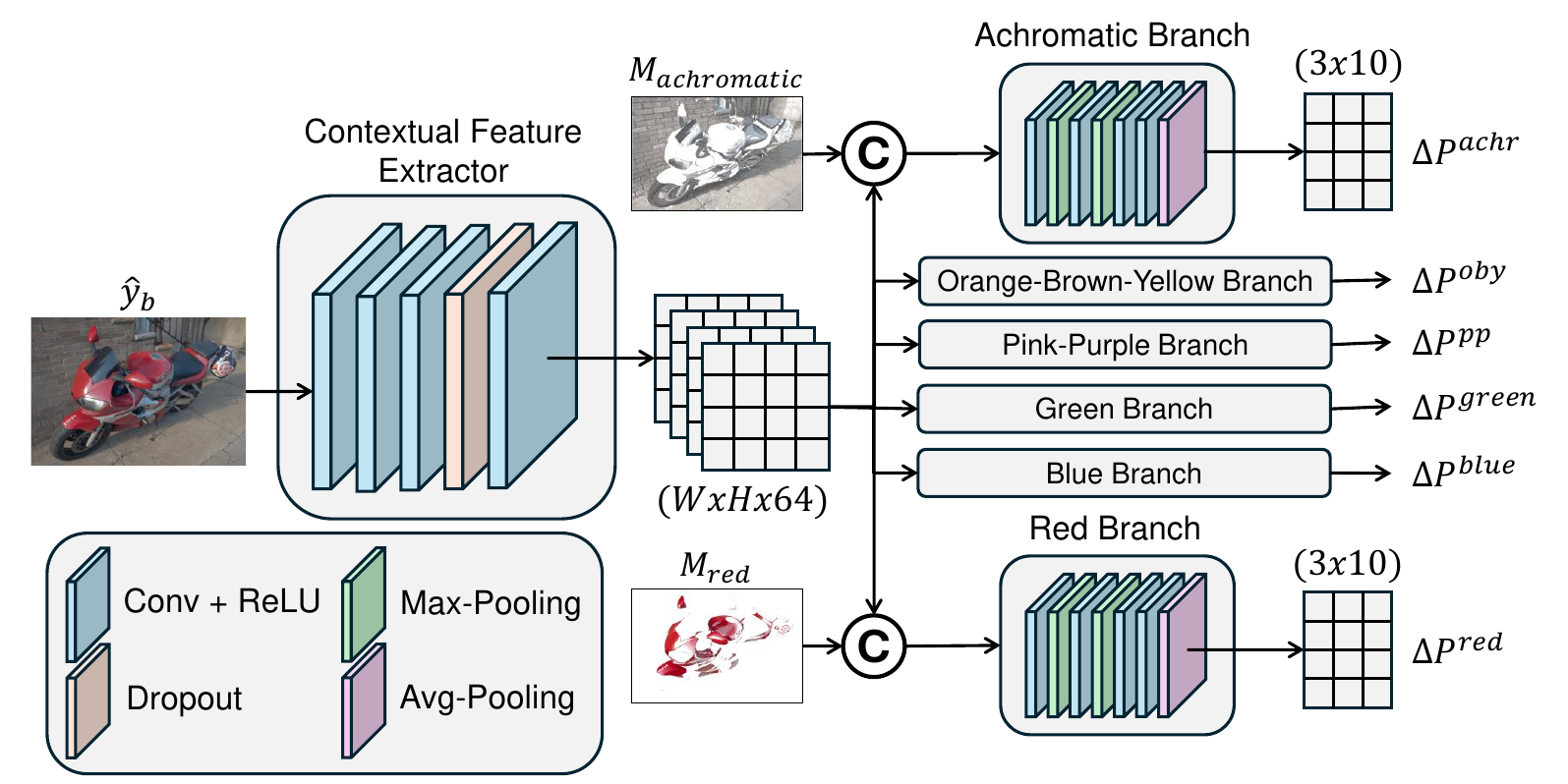}
    \caption{Bezier Control Point Estimator (BCPE). First, we extract 64-D convolutional feature maps from $\hat{y}_b$ using mainly 4 \texttt{Conv-ReLU} blocks. Then, the color naming is concatenated and passed through 4 \texttt{Conv-ReLU-MaxPooling} and a final \texttt{AveragePooling} layer. The output of the module is $\Delta P$, the unnormalized control points increments.}
    \label{fig:bcpe}
\end{figure}

The Bezier Control Points Estimator (BCPE) aims to estimate the control points defining the Bezier tonal curves of an image. Figure \ref{fig:bcpe} illustrates the BCPE, comprising two distinct blocks: the contextual feature extractor and $6$ color naming branches. The contextual feature extractor primarily consists of $4$ \texttt{Conv-ReLU} blocks and a \texttt{Dropout}, while the color naming branches consist of $4$ \texttt{Conv-ReLU-MaxPooling} blocks and a final \texttt{AveragePooling} layer to manage the variable sizes of the images. 

The contextual feature extractor computes a $64$-d convolutional feature map from the standardized image $\hat{y}_b$. Each color naming branch receives as input a concatenation of the $64$-d feature maps and the corresponding color naming probability map. The output of the branch for color name $n$ consists of three sets of $M$ increments $\left(\Delta P^{n,c}_m\right)_{m=1}^{M}$, each set corresponding to a curve for a given color channel $c$. These increments $\Delta P^{n,c}_m$ do not directly represent the control points as we impose two different constraints. First, to make the curves monotonically increasing functions, we define $\Delta P^{n,c}_m$ as positive increments relative to the previous point. Second, we normalize $\Delta P^{n,c}_m$, ensuring the total increment between the first and last points is 1 and, thus, placing the last point at $(1, 1)$. Consequently, we compute the control points $P^{n,c}_m$ as the accumulated sum of the normalized $\Delta P^{n,c}_m$. This can be formulated as:

\begin{equation}
    P_m^{n, c} = \frac{1}{S^{n,c}} \sum_{k=1}^{m} \Delta P_k^{n, c},
    \label{eq:controlpoints}
\end{equation}
where $S^{n,c}= \sum_{k=1}^{M} \Delta P_k^{n, c}$.

\begin{figure}[t!]
    \centering
    \includegraphics[width=\linewidth]{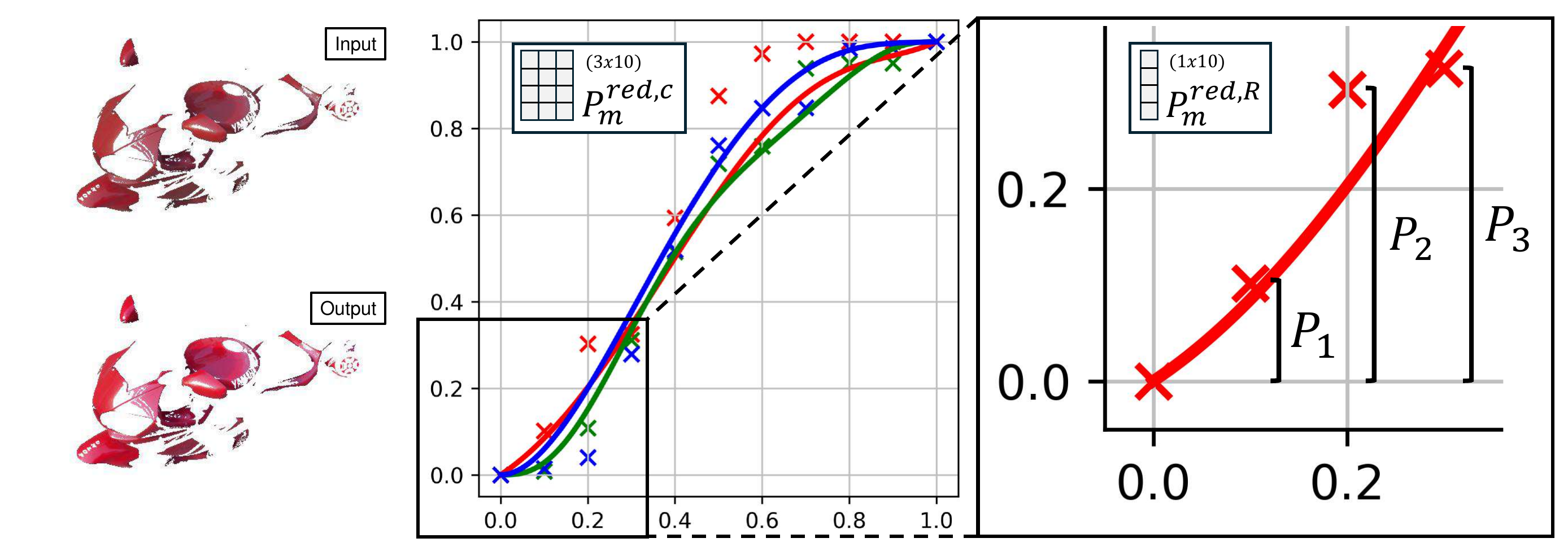}
    \caption{An example of a Bezier curve (Red branch). The first column shows the input and edited image pixels in which the red color name branch focuses. The center plot shows the tone RGB curves learned for the red color name, while the right plot displays a zoomed-in view of the first four control points of the red channel.}
    \label{fig:controlpoints-plot}
\end{figure}

Figure \ref{fig:controlpoints-plot} illustrates an example of a Bezier curve. The left column shows the input and output image pixels with higher red color name probability than 0.2 The center plot shows the tone curves learned for the red color name, while the right plot displays a zoomed-in view of the first four control points of the red channel. Note how the control points are fixed and evenly distributed along the input axis, while $P^{n,c}_{m}$ define the output axis value and, thus, the curvature of the tonal curve. The six sets of Bezier curves learned are applied pixel-wise to the entire standardized image $\hat{y}_b$, yielding six distinct globally-adjusted images.

\subsection{Attention-based image fusion}\label{subsec:imagefusionbyattention}

Tone curves allow global color manipulation but cannot mimic local manipulation from experts in the training images. Here, we detail how to fuse these images via an attention-based mechanism that models their spatial dependencies.

For each of the $6$ color name processed images, we use an attention module to compute a blending weight per image. Specifically, we employ $\hat{y}_b$ as the \textit{Query}, while the corresponding globally-adjusted image serves as the \textit{Key} and \textit{Value}. Initially, we process the input images through 2 \texttt{Conv-ReLU-MaxPooling} blocks, yielding 16 convolutional feature maps with a broad receptive field. In particular, $Q, K, V \in \mathbb{R}^{W/8 \times H/8 \times 16}$. Finally, we aggregate $\hat{y}_b$ using two \texttt{Conv-ReLU- Upsampling} blocks. After the attention mechanism, we aggregate $\hat{y}_b$ via two \texttt{Conv-ReLU-Upsampling} blocks. Note that there is a trade-off between the attention resolution and the computational cost. Downsampling the image by 8 did not introduce noticeable artifacts on the final image $\hat{y}$ after upsampling.

Ultimately, to generate the final image $\hat{y}$, we compute a weighted average of the obtained globally- and locally-adjusted images employing the original color naming probability maps. To mitigate potential artifacts arising from low probability values, we threshold the probability maps at 0.2, setting smaller values to 0. Subsequently, we normalize the maps to ensure each pixel sums to unity before performing the weighted average.

\subsection{Loss Function}\label{subsec:lossfuction}
Our training loss function comprises three terms. The initial term calculates the $L_2$ loss between the standardized image $\hat{y_b}$ and the ground truth $y$, with a weighting factor $\alpha$. The subsequent two terms compute the $L_2$ and $SSIM$ losses between the output of our model $\hat{y}$ and $y$. The primary objective of the first term is to obtain a good enough standardized output by the backbone. The other two terms are designed to assess the fidelity of the final output. In our experiments, we set $\alpha$ to 0.5 (see supplementary for an ablation study on this term). This value was determined to yield optimal performance and allows the image $\hat{y}_b$ to represent the scene colors accurately. The training loss is defined as:
\begin{equation}
    L(\hat{y_b}, \hat{y}, y) = \alpha ||y - \hat{y_b}||_2 + ||y - \hat{y}||_2 + (1 - SSIM(y, \hat{y})).
    \label{eq:loss}
\end{equation}

\section{Experimental Results}
\subsection{Experimental setup}\label{subsec:datasets}

\paragraph{\textbf{Datasets:}}~We compare our method with state-of-the-art (SOTA) methods using the widely used MIT-Adobe-5K dataset \cite{bychkovsky2011learning} and the PPR10K dataset \cite{liang2021ppr10k}. MIT5K consists of 5000 images captured independently using several DSLR cameras and retouched by five artists. However, although the image content is the same, different "versions" have emerged due to variations in image pre-processing and the number of training images. To make a fair comparison with all the SOTA methods, we have used three versions: (1) DPE \cite{chen2018deep}, (2) UPE \cite{wang2019underexposed}, and (3) UEGAN \cite{ni2020towards}. Specifically, DPE splits the data into 2250, 2250, and 500 images for training, validation, and testing, respectively. The last two versions split the images into 4500 images for training and 500 images for testing. DPE and UEGAN pre-process the images in the same manner but with different image sizes, while UPE pre-processes the input images to be under-exposed. Following \cite{moran2021curl, chen2018deep, li2023flexicurve, moran2020deeplpf, wang2019underexposed} we only use the Expert-C retouched images as ground truth. PPR10K is a portrait photo retouching dataset with 11616 high-quality images retouched by 3 experts independently. We use the official splits, dividing the images into 8875 and 2286 for training and testing, respectively. Following \cite{yang2022adaint} we conducted the experiments on the 360p augmented version which every image pair has 5 extra input versions with different manual settings. As in \cite{zeng2020learning, liang2021ppr10k, yang2022adaint} we evaluate our method on the three expert-retouched versions of PPR10K.

\paragraph{\textbf{Implementation Details:}}~We trained our model using Adam~\cite{kingma2014adam}, an initial learning rate 1e-4, reduced by 50\% every 50 epochs. We use horizontal flips for augmenting the training data. We chose the model with the best validation $\Delta E_{00}$ in the DPE version of the MIT5K, while we trained for a fixed 200 and 100 epochs for the other versions of MIT5K and PPR10K, respectively.

\subsection{Comparison with SOTA Methods}\label{subsec:comparisonwithstateoftheartmethods}

We compare our method with SOTA methods using the MIT5K and the 
PPR10K datasets, using the corresponding evaluation metrics. Specifically, in the MIT5K comparisons, we used PSNR, SSIM, LPIPS \cite{zhang2018unreasonable}, $\Delta E_{00}$ and $\Delta E_{ab}$. MIT5K results are presented in Table \ref{tab:MIT5K}. We also report the inference time for a 480p image on an AMD EPYC 7642 and a single NVIDIA A40. Our full architecture outperforms contemporary curve estimation methods; CURL \cite{moran2021curl}, FlexiCurve \cite{li2023flexicurve} and LTMNet \cite{zhao2022learning} on the DPE and UPE versions of MIT5K. Similarly, our method outperforms all-in-one, image-to-image, and LUT-based methods across all the versions. Table \ref{tab:ppr10k} shows results on the PPR10K dataset. Following \cite{liang2021ppr10k, yang2022adaint}, we used PSNR and $\Delta E_{ab}$ to compare our method with the state-of-the-art. As the other methods did not compute SSIM, LPIPS and $\Delta E_{00}$ and the pre-trained models are unavailable, we report these metrics for our model in the supplementary material. We outperform all the contemporary methods on both PSNR and $\Delta E_{ab}$ on the three expert versions of the dataset.

\begin{table}[t!]
  \caption{Quantitative comparisons on the DPE, UPE, and UEGAN versions of the MIT-Adobe-5K dataset. \enquote{-} means the source code or models are unavailable, or results for the corresponding metric were not in the original paper.}
  \label{tab:MIT5K}
  \centering
  \begin{tabular}{clcccccc}
    \toprule
    MIT-5K & \multicolumn{1}{c}{\centering Method} & PSNR $\uparrow$ & SSIM $\uparrow$ & LPIPS $\downarrow$ & $\Delta E_{00}$ $\downarrow$ & $\Delta E_{ab}$ $\downarrow$ & Time (ms)\\
    \midrule
    \multirow{5}{*}{DPE} & DPE \cite{chen2018deep} & 23.80 & 0.900 & - & - & - & - \\
    & DeepLPF \cite{moran2020deeplpf} & 23.93 & 0.903 & 0.040 & 7.00 & 8.05 & 136 \\
    & CURL \cite{moran2021curl} & 24.04 & 0.900 & - & - & - & 102 \\
    & FlexiCurve \cite{li2023flexicurve} & 24.37 & 0.920 & 0.060 & - & - & -\\
    & NamedCurves (Ours) & \textbf{24.91} & \textbf{0.927} & \textbf{0.038} & \textbf{6.60} & \textbf{7.82} & 26 \\
    \midrule
    \multirow{5}{*}{UPE} & DPE \cite{chen2018deep} & 22.15 & 0.850 & 0.108 & - & - & - \\
    & UPE \cite{wang2019underexposed} & 23.04 & 0.893 & 0.158 & - & - & - \\
    & LTMNet \cite{zhao2022learning} & 24.27 & \textbf{0.913} & 0.068 & - & - & - \\
    & DeepLPF \cite{moran2020deeplpf} & 24.48 & 0.887 & 0.103 & 6.89 & 7.77 & 136 \\
    & NamedCurves (Ours) & \textbf{25.20} & 0.906 & \textbf{0.047} & \textbf{6.54} & \textbf{7.58} & 26 \\
    \midrule
    \multirow{5}{*}{UEGAN} & InstructIR \cite{conde2024high} & 24.65 & 0.900 & - & - & 7.61 & - \\
    & 3DLUT \cite{zeng2020learning} & 25.29 & 0.923 & 0.043 & 6.76 & 7.55 & 13 \\
    & SepLUT \cite{yang2022seplut} & 25.47 & 0.921 & 0.042 & 6.71 & 7.49 & 10 \\
    & AdaInt \cite{yang2022adaint} & 25.49 & 0.926 & 0.041 & 6.69 & 7.47 & 13 \\
    & NamedCurves (Ours)  & \textbf{25.59} & \textbf{0.936} & \textbf{0.038} & \textbf{6.07} & \textbf{7.40} & 26 \\
    
  \bottomrule
  \end{tabular}
\end{table}

\begin{figure}[t!]
\begin{center}
    \begin{subfigure}{\linewidth}
    \centering
        \includegraphics[width=0.95\linewidth]{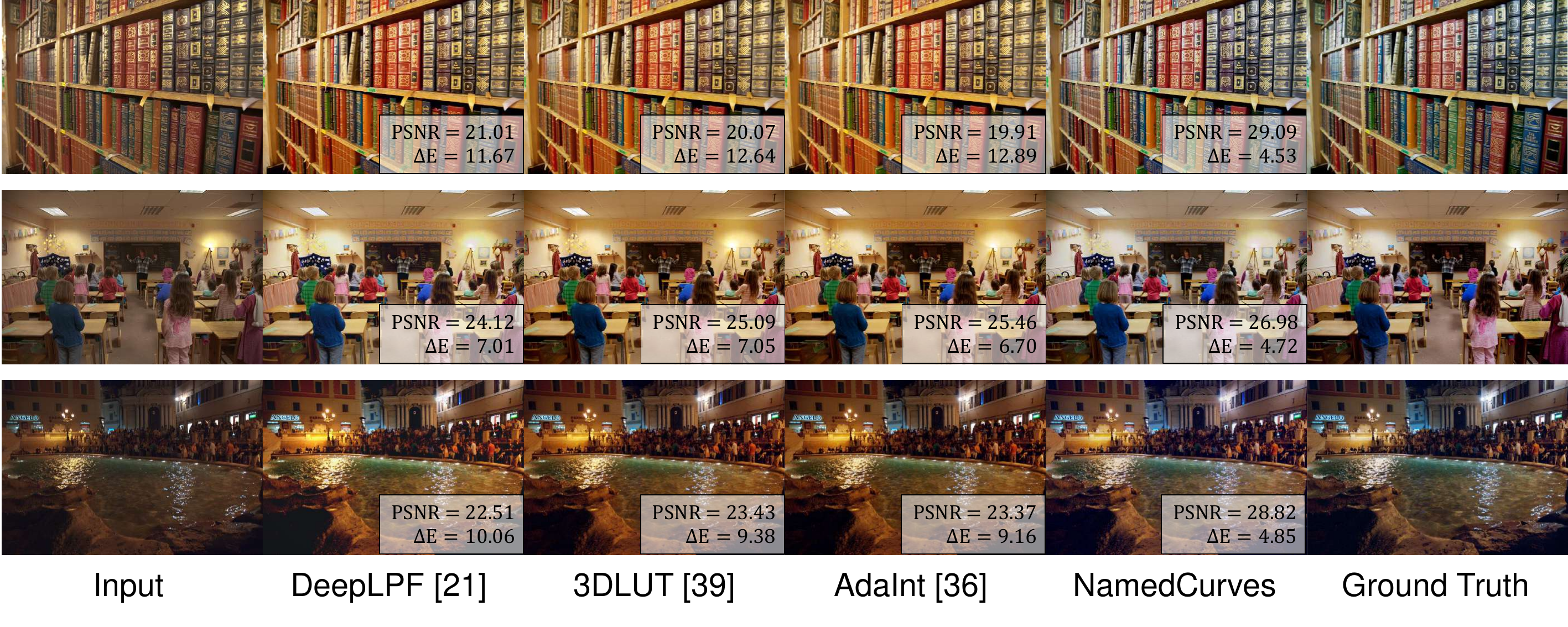}
        \caption{MIT-5K dataset qualitative results.}
        \label{fig:mit5kqualitativeresults}
    \end{subfigure}
    \begin{subfigure}{\linewidth}
        \centering
        \includegraphics[width=0.95\linewidth]{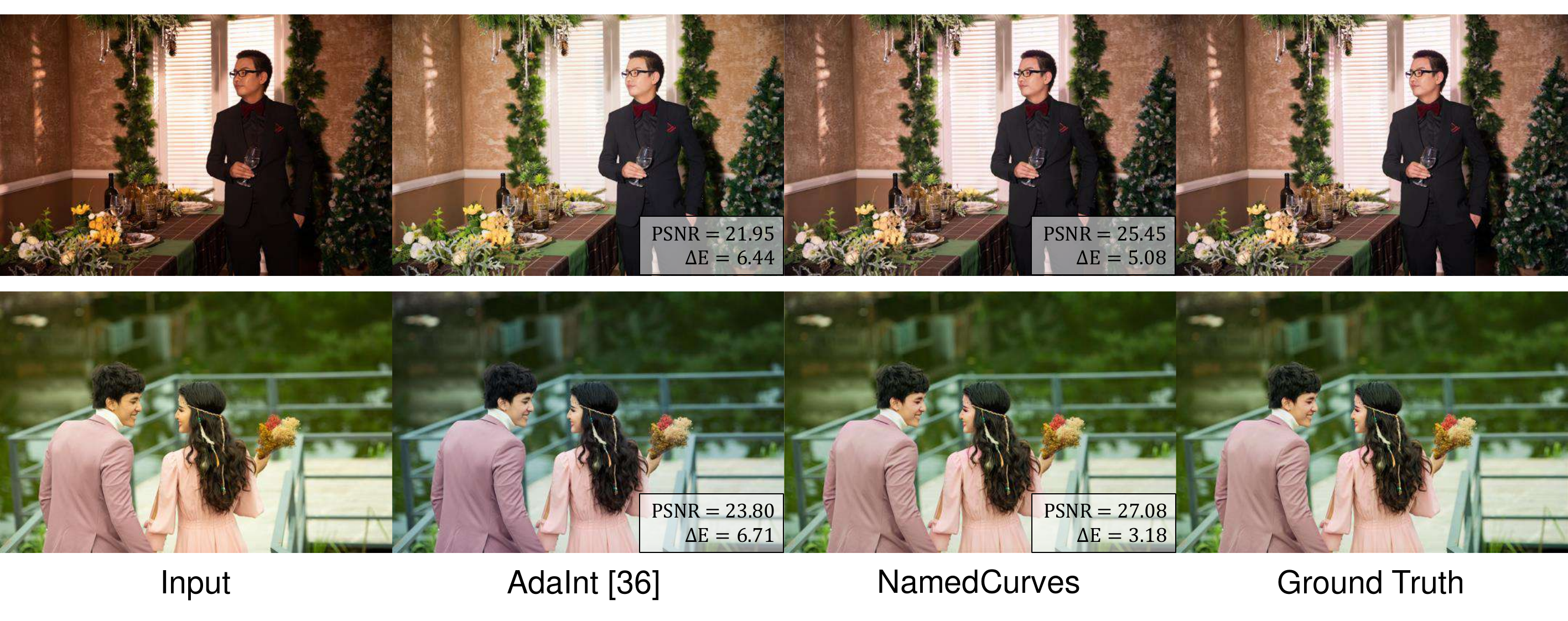}
        \caption{PPR10K dataset qualitative results.}
        \label{fig:PPR10Kqualitativeresults}
    \end{subfigure}
    \caption{Qualitative comparisons on the MIT-5K (Figure \ref{fig:mit5kqualitativeresults}) and PPR10K datasets (Figure \ref{fig:PPR10Kqualitativeresults}). On the bottom-right of each image we display the PSNR and $\Delta E_{00}$.}
    \label{fig:sota1}
    \end{center}
\end{figure}

\begin{table}[t!]
     \caption{Quantitative comparisons the PPR10K dataset. We only compute mean PSNR and $\Delta E_{ab}$ since most other methods do not have models available for inference.}
     \label{tab:ppr10k}
    \centering
    \begin{tabular}{lcccccc}
    \toprule
    \multicolumn{1}{c}{\centering PPR10K} & \multicolumn{2}{c}{ Expert A } & \multicolumn{2}{c}{ Expert B } & \multicolumn{2}{c}{  Expert C } \\
    \cmidrule{2-7}
    \multicolumn{1}{c}{\centering Method} & PSNR $\uparrow$ & $\Delta E_{ab}$ $\downarrow$ & PSNR $\uparrow$ & $\Delta E_{ab}$ $\downarrow$
 & PSNR $\uparrow$ & $\Delta E_{ab}$ $\downarrow$ \\
    \midrule
    HDRNet \cite{li2019hdrnet}              & 23.93 & 8.70 & 23.96 & 8.84 & 24.08 & 8.87 \\
    3DLUT \cite{zeng2020learning}           & 25.64 & 6.97 & 24.70 & 7.71 & 25.18 & 7.58 \\
    SepLUT \cite{yang2022seplut}            & 26.28 & 6.59 & 25.23 & 7.49 & 25.59 & 7.51 \\
    AdaInt \cite{yang2022adaint}            & 26.33 & 6.56 & 25.40 & 7.33 & 25.68 & 7.31 \\
    NamedCurves (Ours)                      & \textbf{26.81} & \textbf{6.48} & \textbf{25.91} & \textbf{7.18} & \textbf{25.69} & \textbf{7.27} \\
    \bottomrule
    \end{tabular}
\end{table}

Figure \ref{fig:sota1} provides several qualitative comparisons, showing examples from both the MIT5K and the PPR10K datasets in Figure \ref{fig:mit5kqualitativeresults} and \ref{fig:PPR10Kqualitativeresults}, respectively. Our method provides visually appealing results that resemble the expert-retouched version regarding color fidelity. In Figure \ref{fig:mit5kqualitativeresults}, for instance, our method accurately rectifies color casts in the first two rows. Lastly, in the third row, our method demonstrates superior performance in enhancing nighttime images. Similarly, in Figure \ref{fig:PPR10Kqualitativeresults} our method outperforms AdaInt \cite{yang2022adaint} in replicating the expert-retouched image on the PPR10K-A dataset. This is particularly noticeable in regions such as the background brown wall (first row) and the global color temperature (last row).

\begin{figure}[t!]
    \centering
    \includegraphics[width=.94\linewidth]{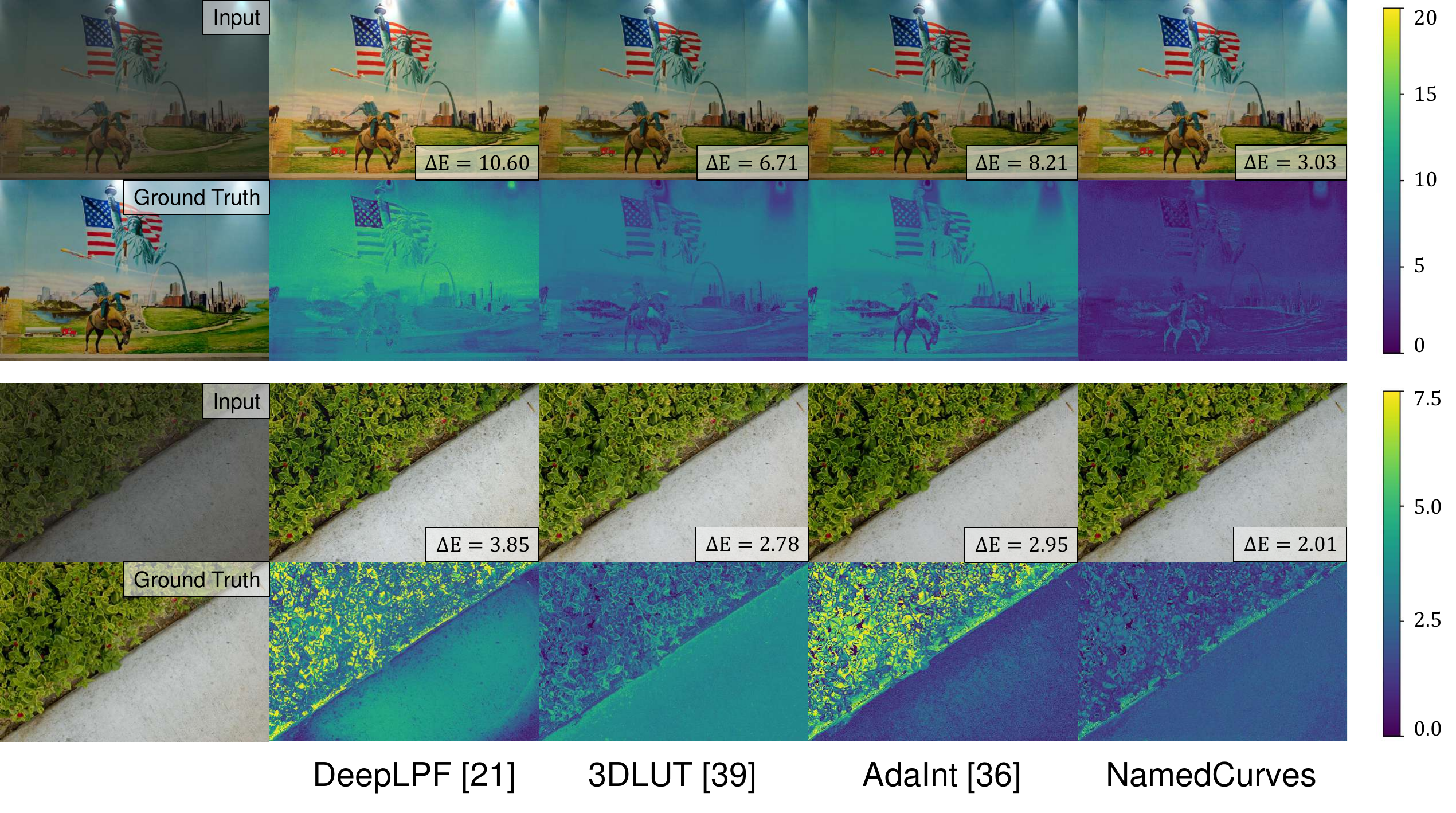}
    \caption{Qualitative results of two images of the MIT-Adobe-5K. The first column presents the input and the expert-retouched image. The top row shows the estimated enhanced version of the input image of DeepLPF \cite{moran2020deeplpf}, 3DLUT \cite{zeng2020learning}, AdaInt \cite{yang2022adaint} and our method. Under each image, we present the $\Delta E_{00}$ error map. On the bottom-right of each image, we display the mean $\Delta E_{00}$.}
    \label{fig:sota2}
\end{figure}

We provide additional qualitative examples in Figure \ref{fig:sota2}. The top row shows the outcomes generated by the methods, while the bottom row displays the per-pixel $\Delta E_{00}$ error maps. In the first row, the other methods struggle to address color cast issues effectively, leading to significant $\Delta E_{00}$ values across the entire image. Conversely, our method demonstrates superior performance, yielding minimal errors confined primarily to small regions. The second image has two distinct areas: one characterized by intricate details and the other by a plain surface. DeepLPF \cite{moran2020deeplpf} exhibits artifacts, such as the elliptical distortion in the grey area. Similarly, 3DLUT \cite{zeng2020learning} and AdaInt~\cite{yang2022adaint} show limitations by enhancing properly only one of the image regions, thereby resulting in substantial errors in the other segment. In contrast, our method consistently enhances the photograph across the entire image by seamlessly integrating both local and global adjustments.

\paragraph{\textbf{User Study:}} We compared our method against AdaInt \cite{yang2022adaint} and SepLUT \cite{yang2022seplut} following a two-alternative forced choice (2AFC), performed in a completely black room with a monitor set to sRGB. We randomly selected 25 images from both MIT5K and PPR10K datasets. 15 observers took part, all tested for colorblindness with the Ishihara test. Results analyzed using Thurstone Case V (larger means better) were: NamedCurves: 1.12; AdaInt: -0.38; SepLUT: -0.74. Our method is statistically significantly better than the other two ---95\% confidence interval is 0.33.

\subsection{Ablation studies}\label{subsec:ablationstudies}

\begin{table}[t!]
    \caption{Ablation studies on the contributions of each module of our proposed method.}
    \label{tab:creditablation}
    \centering
    \begin{tabular}{ccccccccc}
    
    \toprule
    \# & Backbone & Curves & CN & Att. & WN-Avg & PSNR $\uparrow$ & SSIM $\uparrow$ & $\Delta E_{00}$ $\downarrow$\\
    \midrule
    1 &            & 6 & \checkmark & \checkmark & \checkmark & 23.40 & 0.912 & 8.87 \\
    2 & \checkmark &   &            &            &            & 23.74 & 0.916 & 8.68 \\
    3 & \checkmark & 1 &            &            &            & 24.09 & 0.921 & 7.53 \\
    4 & \checkmark & 6 &            &            &            & 24.24 & 0.922 & 7.50 \\
    5 & \checkmark & 6 & \checkmark &            &            & 24.56 & 0.926 & 7.07 \\
    6 & \checkmark & 6 & \checkmark &            & \checkmark & 24.68 & 0.926 & 6.88 \\
    7 & \checkmark & 6 & \checkmark & \checkmark &            & 24.60 & 0.926 & 6.92 \\
    8 & \checkmark & 6 & \checkmark & \checkmark & \checkmark & \textbf{24.91} & \textbf{0.927} & \textbf{6.60} \\
    \bottomrule
    
    \end{tabular}
\end{table}

In this section, we choose the DPE version of MIT5K to conduct several ablation studies to verify the proposed method. We performed experiments to understand the effectiveness of the individual modules used by our framework. Table \ref{tab:creditablation} presents the results for various combinations of the modules of our method. We report PSNR, SSIM, and $\Delta E_{00}$. Throughout the experiments, we consistently utilize the backbone while incrementally incorporating different modules of our method. The column labeled \textit{Curves} indicates the number of RGB Bezier curves utilized and, consequently, the number of globally adjusted images produced. The {\it color naming} column (\textit{CN}) specifies whether the color naming probability maps are concatenated with the 64-d feature maps used by the Beizer curve manipulation. Note that if we use the color naming probability maps we must use six curves. The Attention (\textit{Att.}) column indicates whether we apply local modifications to the globally adjusted images before fusing them. Finally, the \textit{WN-Avg} column denotes whether the experiment employs the color naming maps to weigh the images before blending them. In cases where we do not use WN-Avg and there are multiple output images, we simply average them.

This table shows how combining the different modules of our model improves the performance. Each of our additions improves the result. Experiment 1 emphasizes the importance of the backbone of our model. In detail, color naming modules produce the largest boosts in performance. We gain 0.32 dB in PSNR and 0.43 in $\Delta E_{00}$ when we concatenate the color naming maps to the Bezier Control Point Estimator feature maps - experiment 5. Furthermore, we also gain 0.31 dB in PSNR and 0.32 in $\Delta E_{00}$ when we use the color naming maps to weight the final average - experiment 8.

We further investigate assessing our backbone's impact on our model's performance. We evaluate by using different backbones from other methods. Table \ref{tab:backboneablation} reports the results of our experiments. Notably, our proposed backbone yields superior performance compared to other methods' backbones, namely UNet \cite{moran2020deeplpf}, TED \cite{moran2021curl} and LPIENet \cite{conde2023perceptual}. Importantly, irrespective of the utilized backbone, our method consistently outperforms the previous state-of-the-art models on this DPE version of MIT5K. This highlights the significance of leveraging color naming as the main contributing factor to the performance of our method.

We perform a final ablation study on the number of control points $N$. We tested our method with 5, 7, 11, and 16 control points for each Bezier curve. In Table \ref{tab:controlpointsablation} we report the PSNR, SSIM, and $\Delta E_{00}$ of every experiment. We found that 11 control points --every 0.1 in the input axis--  works best for our method.

\begin{table}[t!]
    \centering
\caption{Ablation studies for the (a) backbone architecture and (b) the number of control points in the Bezier curves.}
\begin{subtable}[t!]{0.47\textwidth}
    \centering
    \begin{tabular}{lccc}
    \toprule
    \multicolumn{1}{c}{\centering Backbone} & PSNR $\uparrow$ & SSIM $\uparrow$ & $\Delta E_{00}$ $\downarrow$\\
    \midrule
    UNet \cite{moran2020deeplpf}        & 24.49 & 0.920 & 7.04 \\
    LPIENet \cite{conde2023perceptual}  & 24.51 & 0.920 & 7.07 \\
    TED \cite{moran2021curl}            & 24.70 & 0.925 & 6.97 \\
    NamedCurves                               & \textbf{24.91} & \textbf{0.927} & \textbf{6.60} \\
    \bottomrule
    \end{tabular} \caption{Ablation on backbone} \label{tab:backboneablation}
\end{subtable}%
\begin{subtable}[t!]{0.43\textwidth}
    \centering
    \centering
    \begin{tabular}{cccc}
    \toprule
    $N$ & PSNR $\uparrow$ & SSIM $\uparrow$ & $\Delta E_{00}$ $\downarrow$\\
    \midrule
    5   & 24.69 & 0.924 & 6.82 \\
    7   & 24.88 & 0.926 & 6.76 \\
    11  & \textbf{24.91} & \textbf{0.927} & \textbf{6.60} \\
    16  & 24.52 & 0.920 & 6.85 \\
    \bottomrule
    \end{tabular} \caption{Ablation on number of control points} \label{tab:controlpointsablation}
\end{subtable}%
\label{tab:table1}
\end{table}

\subsection{Limitations}
\begin{figure}[t]
    \centering
    \includegraphics[width=.95\linewidth]{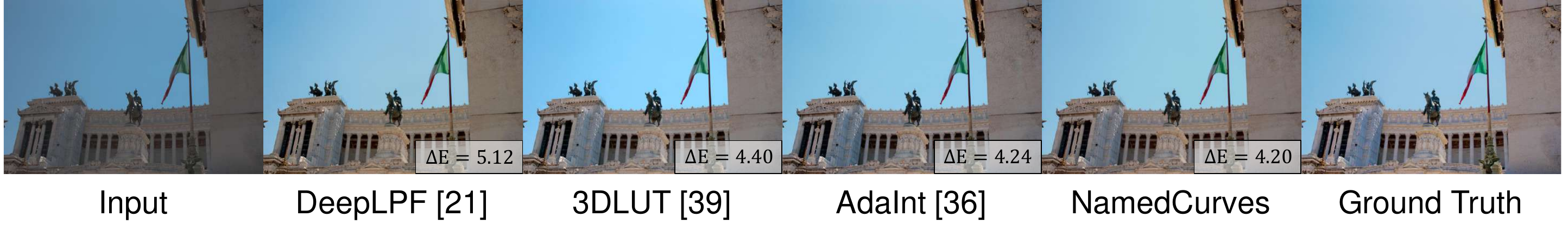}
    \caption{Example of an image with just two dominant colors. Our method still outperforms the others, but does not obtain the same level of advantage.}
    \label{fig:limitations}
\end{figure}

Our method aims to replicate the image style of a skilled photographer estimating a series of tone curves for each color name. However, our method loses part of its advantage compared to the previous SOTA methods in scenarios where the image comprises few color regions. In such instances, different branches of the method receive low-weighting values, resulting in an almost global adjustment technique (e.g. image dominated by just two color names, see Figure~\ref{fig:limitations}).

\section{Conclusion}\label{sec:conclusions}
This paper introduced a new image enhancement model based on color naming that outperforms the current state-of-the-art across various versions of the Adobe 5K and the PPR10K datasets. Our approach uses expert-edited images for learning and explicitly separates the image into a small set of named colors. It learns to adjust the image for each specific named color and then combines the images using an attention-based fusion mechanism. 

\section*{Acknowledgements}
DSL, LH, and JVC were supported by Grant PID2021-128178OB-I00 funded by MCIN/AEI/10.13039/ 501100011033 and by ERDF "A way of making Europe", by the  Departament de Recerca i Universitats from Generalitat de Catalunya with reference 2021SGR01499, and by the Generalitat de Catalunya CERCA Program. DSL also acknowledges the FPI grant from Spanish Ministry of Science and Innovation (PRE2022-101525). LH was also supported by the Ramon y Cajal grant RYC2019-027020-I. MSB was supported by CFREF (VISTA) program, an NSERC Discovery Grant, and the Canada Research Chair program.

\bibliographystyle{splncs04}
\bibliography{main}

\newpage
\section*{Supplementary Material}
The supplemental material provides additional information that could not be incorporated into the main paper due to page limit constraints.  Specifically, we discuss (1) Adobe's color decomposition method, (2) more information on color naming probability maps, (3) further justifications on the color names grouping, (4) loss function parameter $\alpha$ ablation study, and (5) additional results.

\subsection*{Adobe Color Decomposition}
\begin{figure}[t!]
    \centering
    \includegraphics[width=\linewidth]{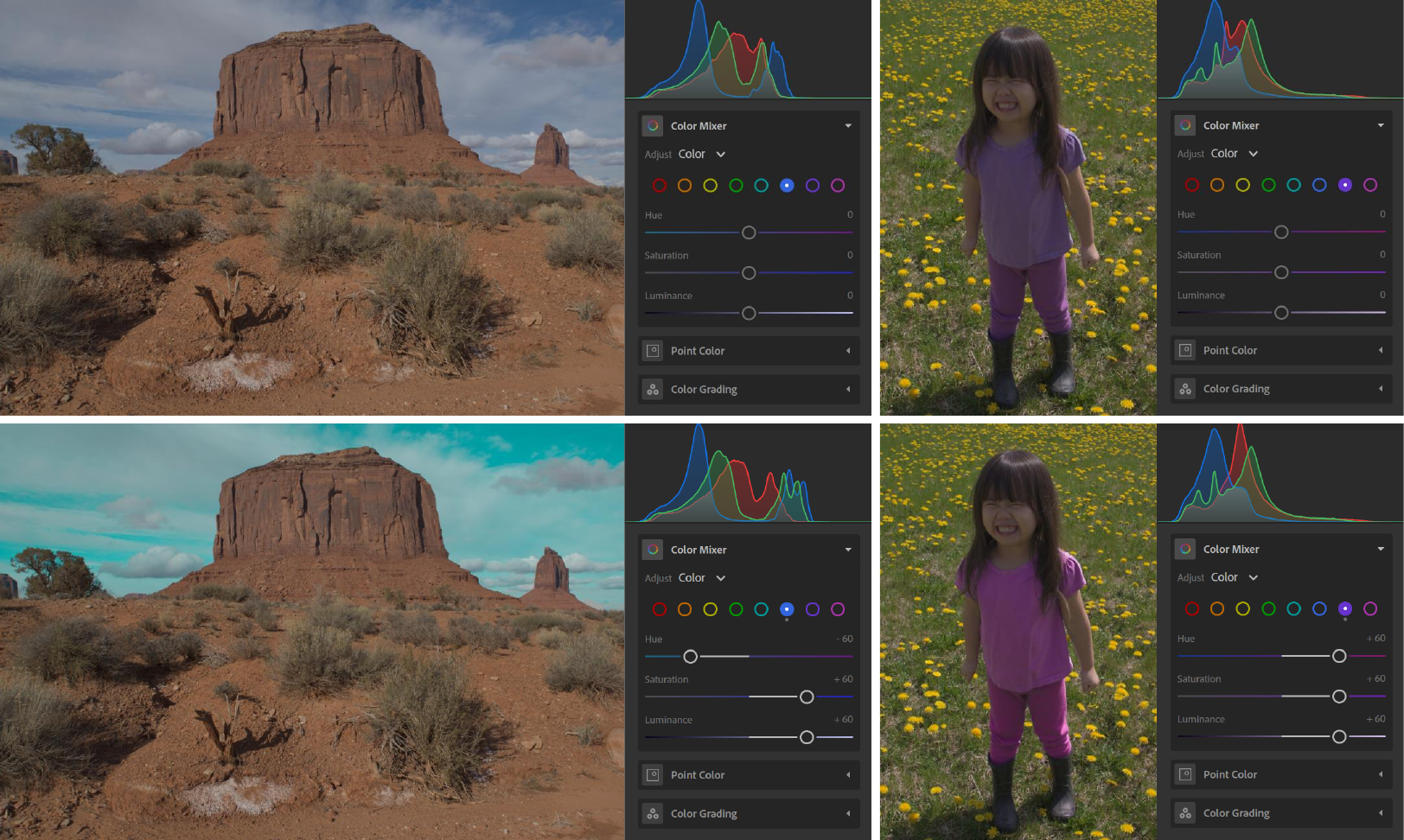}
    \caption{Two examples of the Adobe Color Decomposition tool. In the left example, we manipulate the hue, saturation, and luminance of the color \textit{blue}, while in the right example, we modify the color \textit{purple}.}
    \label{fig:lightroom-decomposition}
\end{figure}

Adobe Photoshop and Adobe Lightroom are software tools to allow photo editors the ability to fine-tune individual colors within an image. Our method is inspired by these tools. In particular, the software decomposes the image into a predefined set of colors (\textit{red}, \textit{orange}, \textit{yellow}, \textit{green}, \textit{cyan}, \textit{blue}, \textit{purple} and \textit{pink}), enabling users to independently manipulate the hue, saturation, and luminance of each color. In Figure \ref{fig:lightroom-decomposition},  we show two screenshots of the tools and examples edited using this feature. For each example, we display the top of the input image alongside the default parameter values for the color to fine-tune. We show the edited image in the bottom images alongside the corresponding slider adjustments. In the left example, we demonstrate the modification of \textit{blue}, illustrating alterations in the sky while preserving non-\textit{blue} regions. In the right example, we focus on adjusting \textit{purple}, a non-primary color. This allows us to selectively modify specific \textit{purple} elements, such as the girl's clothing, without affecting the rest of the image. Notably, adjustments to the desired color sliders induce changes across all three color channels, as evident in the histograms provided in the top-right corner. 

\subsection*{Color Naming Probability Maps}
As discussed in the main paper, we use the color naming method proposed by Van de Weijer et al.~\cite{van2009learning}.  This method is applied on each pixel. Given an sRGB image, this method generates 11 probability maps, each corresponding to a distinct color name: \textit{red, blue, green, yellow, pink, purple, orange, brown, white, grey, black}. As discussed in the main paper, we group certain color names due to their similar hues, differing primarily in intensity only. Specifically, we merge \textit{orange-brown-yellow}, \textit{pink-purple}, and \textit{white-grey-black} (referring to this last one as \textit{achromatic}). The combined maps are obtained by summing the individual probability maps. 
In the end, we obtain probability maps for 6 color categories.

\begin{figure}[t!]
    \centering
    \includegraphics[width=\linewidth]{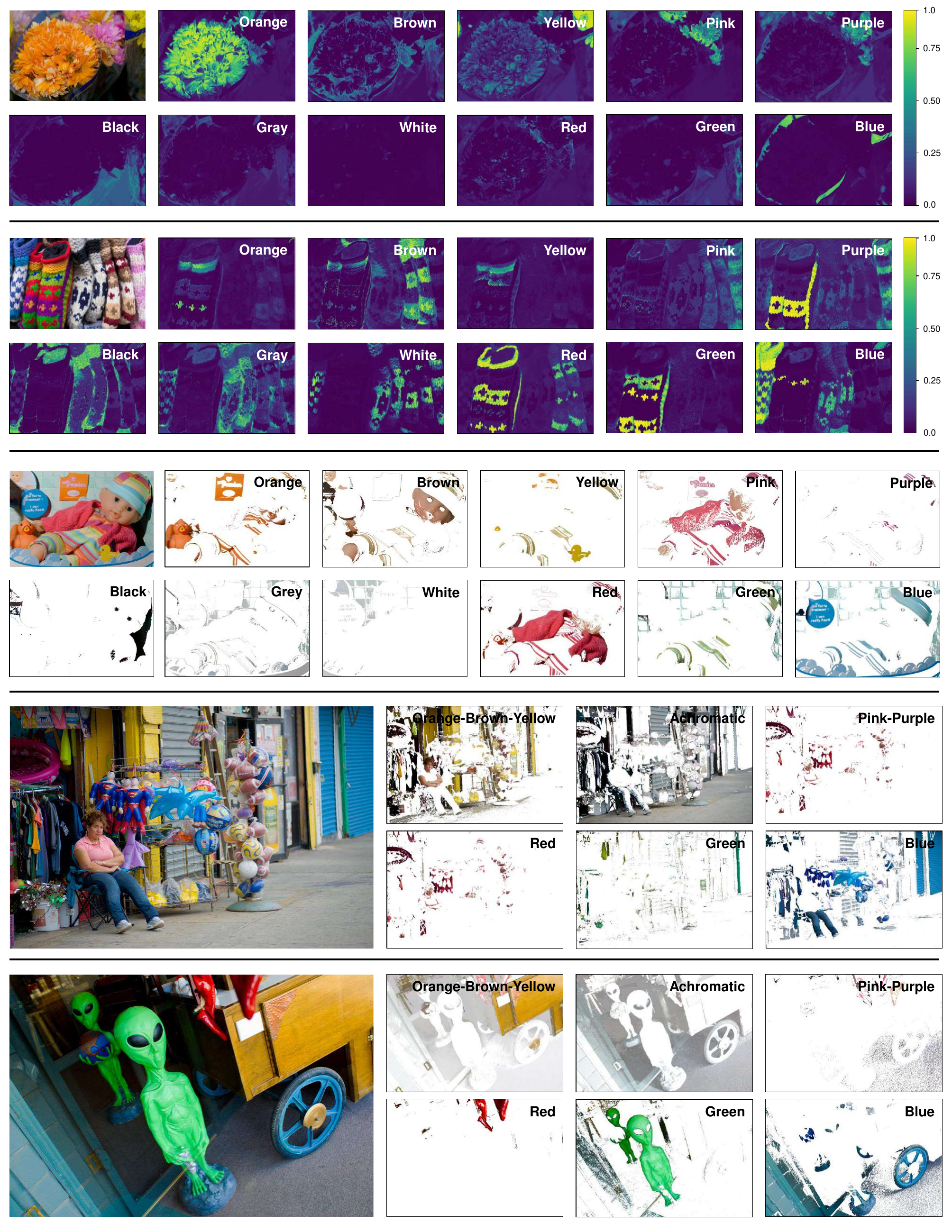}
    \caption{Van de Weijer et al.~\cite{van2009learning} color naming method applied pixel-wise to five images. The color naming probability maps of the first two examples are displayed with the same color map. The third and fourth rows present visualization examples where we display the image pixels whose color naming map is higher than 0.2. The last row presents the same visualization as in the main paper, where we also take into consideration the probability value.}
    \label{fig:color-naming}
\end{figure}

Figure \ref{fig:color-naming} illustrates these color-naming probability maps using different visualizations. In the first two examples, we depict the 11 color naming probability maps using the same color map for all the color names. In the third and fourth examples, we show the image pixels exceeding 0.2 probability for the 11 color names and the 6 color-category version, respectively. In this case, the pixels in the maps represent the real sRGB values in the original image. Finally, in the last example, we use the same visualization method as in the main submission (i.e., Figures 1, 3, and Figure 4).  We also account for the probabilities assigned to each color to emphasize the probability aspect of color naming. In particular, for each pixel and color name, we compute:
\begin{equation}
    n = (1-p_i) I_w + p_i I_i,
    \label{eq:colorcoding}
\end{equation}
where $p_i$ represents the color name probability of pixel $i$, $I_w$ denotes a white RGB value (i.e., [1, 1, 1]) and $I_i$ signifies the RGB value of pixel $i$. It is important to note that the colors represented in these maps are not the original RGB values of the image, as they are scaled by the color naming probability $p$. 

\subsection*{Color Names Grouping}
\begin{figure}[t]
\begin{tabular}{ccc}
    \includegraphics[width=0.32\linewidth]{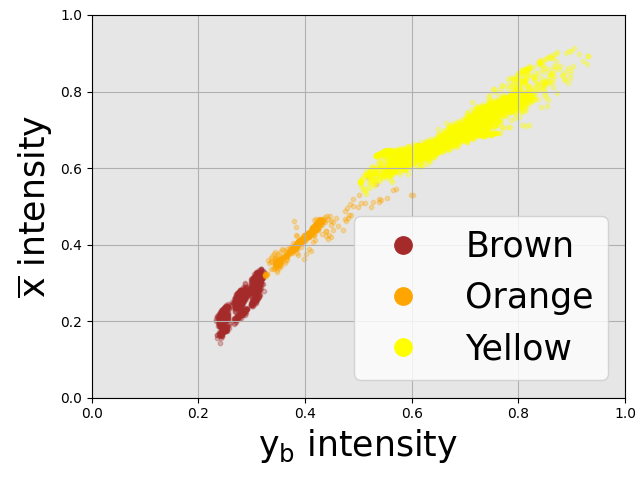}&
    \includegraphics[width=0.32\linewidth]{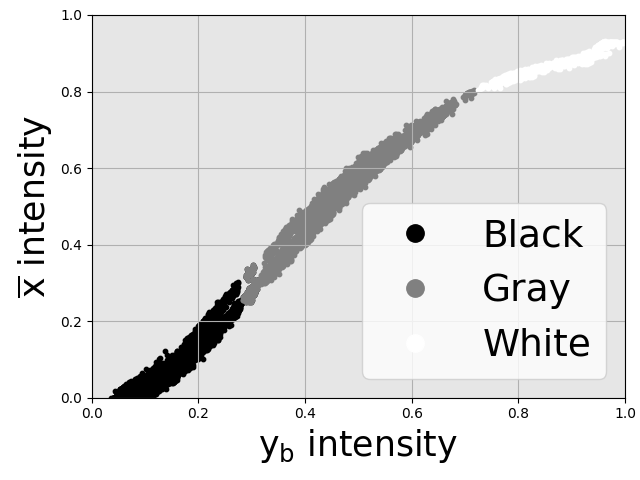}&
    \includegraphics[width=0.32\linewidth]{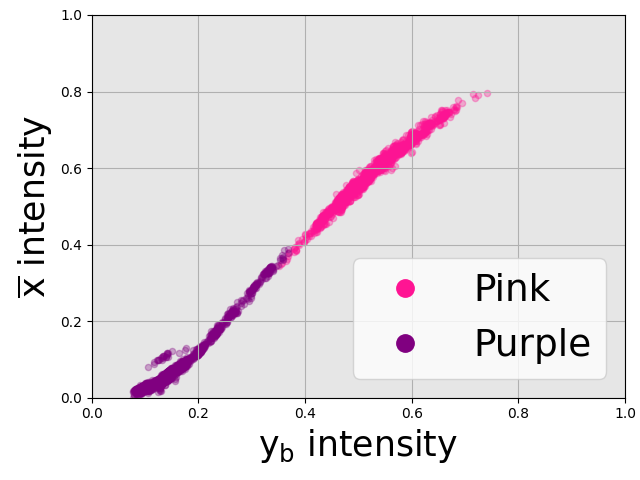} \\
\end{tabular}
\caption{Joined color names with respect to the intensity value. Pixels with probability $>$ 0.5 are plotted.}\label{fig:blending}
\end{figure}

We used the color naming model from Van de Weijer et al. \cite{van2009learning} to obtain the probability map for each color name, namely \textit{red, blue, green, yellow, pink, purple, orange, brown, white, grey, black}. However, we note that some linguistic color names share similar hues, but only differ in intensity. As tone curves are defined for all the intensity ranges, it will be beneficial to group these colors together. To this end, we reduce the set of $11$ probability maps to just $6$ by grouping \textit{orange-brown-yellow}, \textit{pink-purple}, and \textit{white-grey-black} (referring to this last one as \textit{achromatic}). In Figure~\ref{fig:blending}, we visually show the reason for reducing the number of Color Naming channels to just 6. The figure illustrates 2D plots depicting the relationship between input and output intensity values for pixels with probability $>0.5$ belonging to each specific color. For example, in the case of orange-brown-yellow, brown is only present at low intensities, while orange dominates at mid-intensities and yellow at top intensities. The same analysis extends to the other joined color channels. Our method aims to learn a curve to be applied at all the intensity levels. Thus, incorporating information spanning all the intensity levels is beneficial. To also show this numerically, we experimented using our model with the $11$ color terms, obtaining a PSNR of 24.72 dB (in comparison to 24.91 dB with $6$ channels) in the MIT5K-DPE dataset.

\subsection*{Loss Function Ablation Study}

Following prior works (DeepLPF~\cite{moran2020deeplpf}, CURL~\cite{moran2021curl}) our method starts with a backbone that serves to standardize the input. We consider three loss terms: the $l_2$ loss between the output of the backbone and the reference image and the $l_2$ and SSIM losses between the final output and the reference image. We chose to weight the backbone loss by $\alpha=0.5$. An ablation study is shown in Table \ref{tab:alpha_ablation}, where we can see that $\alpha=0.5$  gives better results than $\alpha=0$ (i.e., ignore the backbone output) and $\alpha=1$ (i.e., heavily weight backbone output).

\begin{table}[t]
    \centering
    \caption{Ablation study on the $\alpha$ parameter of the loss function on the MIT5K-DPE dataset.}
    \begin{tabular}{ccc}
    \toprule
    $\alpha$ & PSNR & $\Delta E_{00}$ \\
    \midrule
    0 & 24.58 & 6.76 \\
    0.5 & \textbf{24.91} & \textbf{6.60} \\
    1 & 24.73& 6.62 \\
    \bottomrule
    \end{tabular}

    \label{tab:alpha_ablation}
\end{table}

 \begin{table}[t]
     \caption{Additional quantitative results of our method on the PPR10K dataset.  No other method computes these metrics in their respective papers for the PPR10K dataset.}
     \label{tab:ppr10k1}
     \centering
     \begin{tabular}{ccccc}
     \toprule
     Expert & SSIM $\uparrow$ & LPIPS $\downarrow$  & $\Delta E_{00}$ $\downarrow$ \\
     \midrule
     A & 0.957 & 0.031 & 5.46 \\
     B & 0.956 & 0.032 & 5.61 \\
     C & 0.949 & 0.032 & 5.68 \\
     \bottomrule

     \end{tabular}
 \end{table}

\subsection*{Additional MIT5K and PPR10K qualitative results}
\begin{figure}[t]
    \centering
    \includegraphics[width=.98\linewidth]{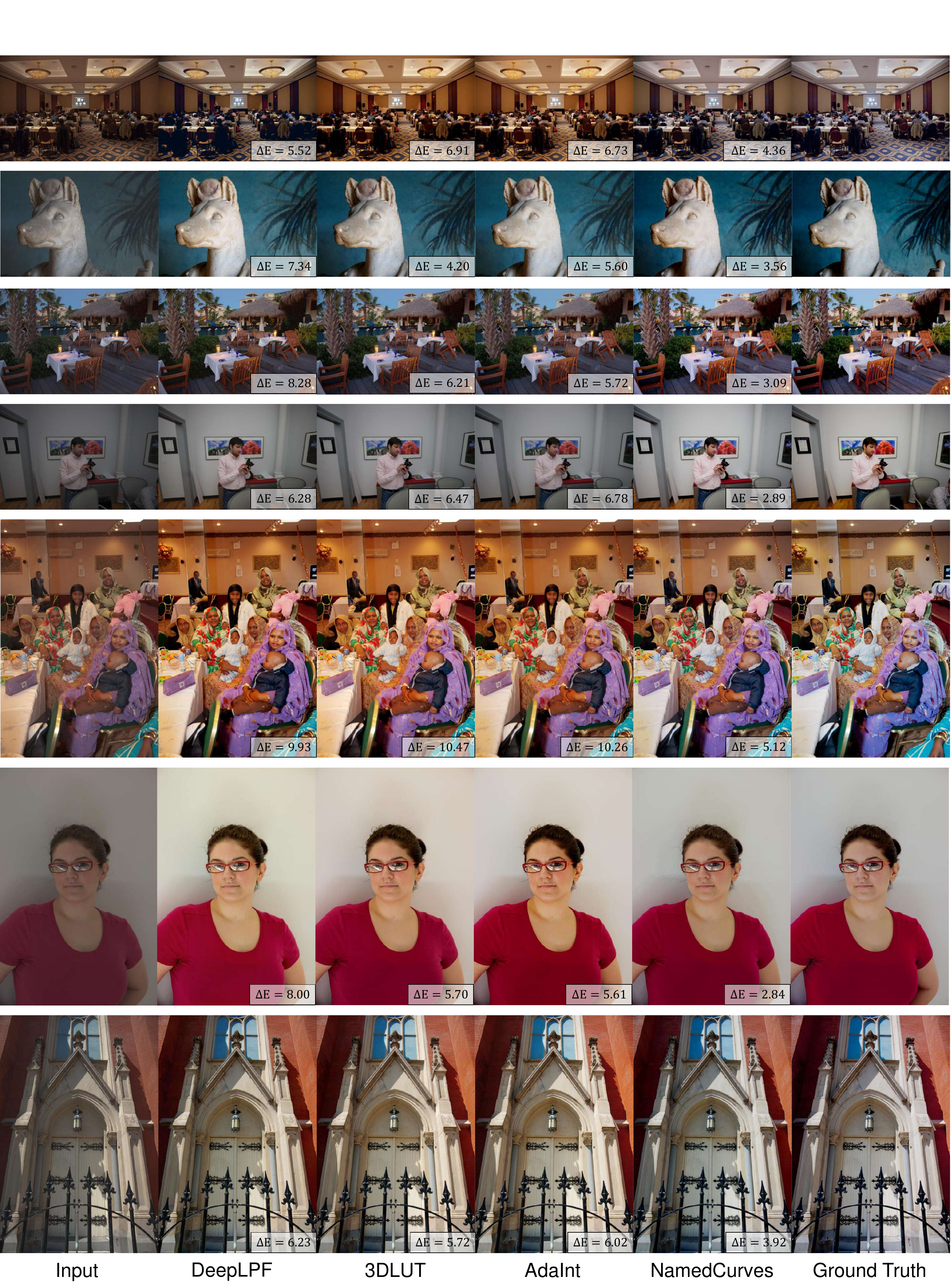}
    \vspace{-1mm}
    \caption{Additional qualitative results on the MIT5K dataset. From left to right: the input image, DeepLPF~\cite{moran2020deeplpf}, 3DLUT~\cite{zeng2020learning}, AdaInt~\cite{yang2022adaint}, our method, and the ground truth. $\Delta E_{00}$ is shown in the bottom-right corner of each image.}
    \label{fig:qualitative-mit5k}
\end{figure}

\begin{figure}[t]
    \centering
    \includegraphics[width=.98\linewidth]{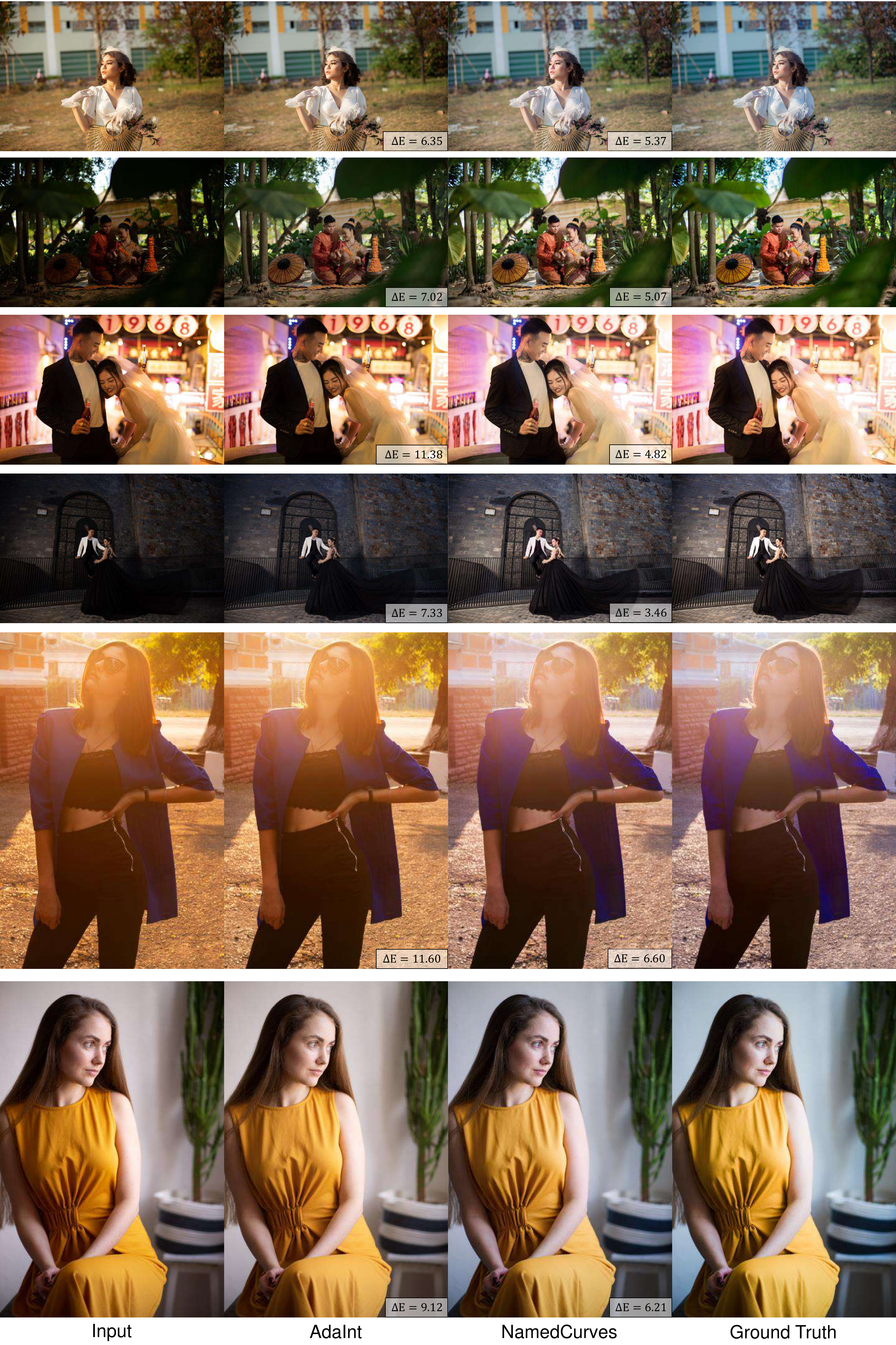}
    \vspace{-2mm}
    \caption{Additional qualitative results performed using the PPR10K dataset. From left to right: the input image, AdaInt~\cite{yang2022adaint}, our method, and the ground truth. $\Delta E_{00}$ is shown in the bottom-right corner of each image.}
    \label{fig:qualitative-ppr10k}
\end{figure}

Figure \ref{fig:qualitative-mit5k} and Figure \ref{fig:qualitative-ppr10k} show additional results from the MIT5K and PPR10K, respectively. We compare our method with DeepLPF~\cite{moran2020deeplpf}, 3DLUT~\cite{zeng2020learning} and AdaInt~\cite{yang2022adaint}.

\subsection*{Additional PPR10K quantitative results}
Table~\ref{tab:ppr10k1} reports SSIM, LPIPS and $\Delta E_{00}$ of our model on experts A, B and C of PPR10K. These metrics are not computed by other methods in their respective papers.

%
%

\end{document}